\documentclass[11pt,a4paper,fleqn]{extarticle}

\usepackage[
    top=20mm,
    bottom=20mm,
    left=18mm,
    right=18mm
]{geometry}

\usepackage{graphicx}
\usepackage{titlesec}
\usepackage{fancyhdr}
\usepackage{array}
\usepackage{booktabs}
\usepackage{tikz}
\usepackage[most]{tcolorbox}
\usepackage{amssymb}
\usepackage{amsmath}
\usepackage{fontspec}
\usepackage{xcolor}
\usepackage[normalem]{ulem}
\usepackage{environ}
\usepackage{etoolbox}

\usepackage[numbers,compress]{natbib}
\usepackage{nicefrac}
\usepackage{microtype}
\usepackage{adjustbox}
\usepackage{float}
\usepackage{placeins}
\usepackage{subcaption}
\usepackage{multirow}
\usepackage{makecell}
\usepackage{enumitem}
\usepackage{xspace}
\usepackage{algorithm}
\usepackage{algpseudocode}
\usepackage{listings}
\usepackage{etoc}
\usepackage{caption}
\usepackage{fvextra}
\usepackage{hyperref}
\usepackage{url}

\titleformat{\section}
    {\sffamily\large\bfseries}
    {\thesection}
    {1em}
    {}

\titleformat{\subsection}
  {\sffamily\normalsize\bfseries}
  {\thesubsection}
  {1em}
  {}

\titleformat{\subsubsection}
  {\sffamily\normalsize\bfseries}
  {\thesubsubsection}
  {1em}
  {}

\titleformat{\paragraph}[runin]
  {\normalsize\bfseries}
  {\theparagraph}
  {1em}
  {}

\etocsettocstyle{\section*{\contentsname}\setlength{\parskip}{0pt}}{}
\makeatletter
\renewcommand*\l@subsection{\@dottedtocline{2}{1.5em}{3.2em}}
\makeatother

\definecolor{tether-gray}{HTML}{F2F1EF}
\definecolor{tether-link}{HTML}{1A7F64}

\hypersetup{
    colorlinks=true,
    linkcolor=tether-link,
    citecolor=tether-link,
    urlcolor=tether-link
}

\newtcolorbox[auto counter]{custombox}[2][]{
    title={Box~\thetcbcounter: #2},
    label={#1},
    fonttitle=\bfseries,
    coltitle=black,
    before title=\vspace{15pt},
    colback=tether-gray,
    colframe=tether-gray,
    boxrule=0pt,
    arc=5pt,
    left=15pt,
    right=15pt,
    bottom=15pt,
    width=\textwidth,
}

\tcbset{tethertable/.style={
    colback=tether-gray,
    colframe=tether-gray,
    boxrule=0pt,
    arc=5pt,
    left=10pt,
    right=10pt,
    top=8pt,
    bottom=8pt,
    before skip=0pt,
    after skip=0pt
}}
\let\tetherorigtable\table
\let\endtetherorigtable\endtable
\renewenvironment{table}[1][tbp]
  {\tetherorigtable[#1]\begin{tcolorbox}[tethertable]}
  {\end{tcolorbox}\endtetherorigtable}

\NewEnviron{ack}{%
  \section*{Acknowledgments and Disclosure of Funding}
  \BODY
}

\newcommand{\paperlink}[3]{%
  \href{#2}{\raisebox{-0.2em}{\includegraphics[height=1.15em]{#1}}\hspace{0.4em}\textcolor{black}{\bfseries #3}}%
}

\newtcolorbox{promptbox}[1][]{%
  enhanced,
  breakable,
  colback=tether-gray,
  colframe=tether-gray,
  colbacktitle=tether-gray!88!black,
  coltitle=black,
  boxrule=0pt,
  arc=5pt,
  fonttitle=\sffamily\bfseries\small,
  left=10pt, right=10pt, top=6pt, bottom=6pt,
  #1%
}
\DefineVerbatimEnvironment{promptverbatim}{Verbatim}{%
  fontsize=\small,
  breaklines=true,
  breakanywhere=true,
  breaksymbolleft={},
  breakindent=0pt,
  formatcom=\raggedright%
}

\newcommand{\papertitle}{QVAC Genesis III: A Large-Scale, High-Quality Open Synthetic STEM Corpus for Efficient Language Model Pre-Training}

\newcommand{\paperauthors}{%
Davide Vitabile,
N. Ranjan,
Akshay Nambiar,
Kamal K. Gupta,
Amril Nazir%
}

\newcommand{\paperaffiliation}{Tether Data, S.A. de C.V. d.b.a. Tether AI Research}

\newcommand{\huggingfaceurl}{https://huggingface.co/datasets/qvac/GenesisIII}
\newcommand{\githuburl}{https://github.com/tether-ai-research/qvac-genesis-III}

\newcommand{\paperabstract}{%
High‑quality pre‑training data is a critical bottleneck for educational and STEM‑specific language models targeting edge AI and on‑device deployment where token budgets are tightly constrained. While major organizations train ever-larger models on private corpora, the open ecosystem lacks STEM-focused synthetic datasets that deliver high per-token learning value efficiently for small models. To address this gap, we introduce QVAC Genesis III, a 191.43B-token, STEM-focused multi-domain synthetic corpus covering 19 domains across several difficulty levels and different educational styles. QVAC Genesis III is built via a dual generation strategy that performs targeted teacher distillation using a weak edge-scale student model as signal: the student's \emph{failures} are converted into corrective explanations, while its \emph{successes} are expanded into contrastive option-level reasoning over all answer choices. We further introduce an \emph{LLM-as-a-parser} evaluation protocol that extracts final answers from free-form outputs and tracks both accuracy and answer validity. To validate the effectiveness of our QVAC Genesis III data, we conduct controlled from-scratch ablations with 1.7B-parameter models, showing that models trained with QVAC Genesis III consistently outperform both models trained with the open-source synthetic corpus Cosmopedia-v2 and the publicly released Cosmo-1B model across ARC, GPQA Diamond, and MMLU STEM benchmarks, achieving up to +28.57\% on ARC-E and +21.35\% on ARC-C, while reaching a Valid Answer Rate of up to 99.45\%.%
}

\begin{document}

\noindent
\includegraphics[height=20pt]{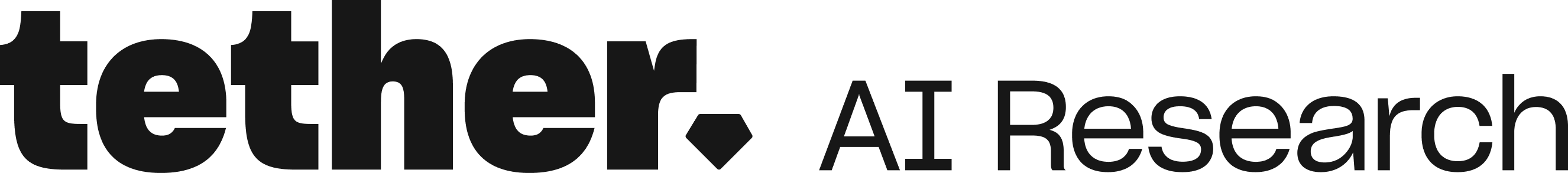}%
\hfill
\raisebox{0.5pt}{\includegraphics[height=14pt]{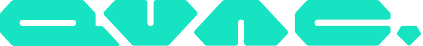}}
\par\medskip

\vspace{10pt}

\begin{tcolorbox}[
    colback=tether-gray,
    colframe=tether-gray,
    boxrule=0pt,
    arc=5pt,
    left=15pt,
    right=15pt,
    top=15pt,
    bottom=15pt,
    width=\textwidth
]

{\sffamily\fontsize{20.74}{25}\selectfont\bfseries\raggedright\hyphenpenalty=10000
\papertitle
\par}

\vspace{5pt}

{\sffamily\fontsize{11}{13.6}\selectfont\bfseries
\paperauthors
\par}

\vspace{2.5pt}

{\sffamily\fontsize{11}{13.6}\selectfont
\paperaffiliation
\par}

\vspace{6pt}

{\fontsize{11}{13.6}\selectfont
\sffamily
\paperlink{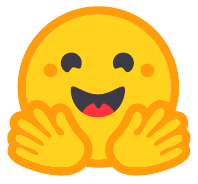}{\huggingfaceurl}{\mdseries\huggingfaceurl}\\[3pt]
\paperlink{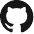}{\githuburl}{\mdseries\githuburl}
\par}

\vspace{15pt}

{\normalsize
\paperabstract
\par}

\end{tcolorbox}

\suppressfloats[t]

\vspace{10pt}

\section{Introduction}
\label{sec:intro}

Educational and STEM-specific (Science, Technology, Engineering, and Mathematics) language models are essential because they push large language models (LLMs) beyond basic text generation toward precise reasoning and complex problem solving. Incorporating STEM-focused data directly improves factual accuracy and logical robustness. Yet, despite this need, high-quality STEM pre-training data remains scarce in the open ecosystem. The primary source, large-scale web crawls, is underrepresented in STEM domains and lacks rigorous correctness verification and multi-domain categorization, making it unreliable for education and professional settings that require factual precision and disciplined reasoning. Moreover, major organizations maintain carefully curated STEM corpora that are not publicly released, further widening the gap between well-resourced labs that train models at massive scale and academic or smaller teams aiming to build efficient models for practical deployment. For customized models that must remain computationally feasible on modest compute infrastructure during training, pre-training under tightly constrained token budgets is critical because each training token must deliver maximum learning value \citep{hoffmann2022training}. There is also a growing need to run these models on edge devices (smartphones, laptops and embedded systems) with limited internet connectivity and on privacy-sensitive on-premise servers that must operate in offline mode. This has created strong demand for small, efficient models in the 1--2B parameter range that can operate under strict memory, latency, and energy constraints. 

Synthetic data has emerged as a key strategy to close this gap. Microsoft's Phi series demonstrated that large-scale synthetic datasets can train competitive small models~\citep{li2023phi15}, and HuggingFace's Cosmopedia provided the first open replication of this approach~\citep{benallal2024cosmopedia}. However, existing open synthetic corpora remain insufficiently STEM-focused and lack the \emph{token efficiency} needed for reasoning-intensive pre-training at edge scale. When parameter counts and token budgets are both constrained, the \emph{structure} and \emph{quality} of training data matter far more than sheer volume.

A further inefficiency is that current pipelines systematically discard the learning signal embedded in model failures. Incorrect answers are typically filtered out to avoid contaminating the corpus, but this also removes examples that expose conceptual gaps and misconceptions, high-value supervision that could improve robustness and reasoning, especially for small models that cannot afford to waste capacity on redundant content.

We address these limitations with \textbf{QVAC Genesis III}, a \textbf{191.43B-token}, STEM-focused, multi-domain educational synthetic corpus spanning \textbf{19 domains} and \textbf{three difficulty levels} (high-school, college, and professional). QVAC Genesis III is built via a \emph{dual generation strategy} (Figure~\ref{fig:dual_pipeline}) that performs \emph{targeted teacher distillation} from a capable reasoning model into edge-scale student models: the student's \emph{failures} and \emph{successes} are both converted into structured, high-value training content:
\begin{enumerate}
    \item \textbf{Failure Analysis (FA) pipeline.} When the student produces an incorrect answer, we trigger a teacher-driven procedure that diagnoses the likely misconception and generates a targeted corrective explanation, turning each failure into a focused teaching moment.
    
    \item \textbf{Option-Level (OL) reasoning pipeline.} When the student succeeds, the teacher expands the instance into detailed \textit{option-level reasoning} over \emph{all} answer choices: explaining why the correct option holds and why each distractor fails, yielding richer supervision than single-trace solutions.
\end{enumerate}

Together, these pipelines ensure that no student signal is wasted: failures become corrective lessons, and successes become comprehensive reasoning demonstrations. Content is rendered in four complementary styles (educational textbook, web articles, question-answering, and conversational dialogue) to maximize diversity. We further introduce an \emph{LLM-as-a-parser} evaluation protocol (Figure~\ref{fig:llm_extractor}) that extracts final answers from full model-generated output and reports both accuracy and a \textit{Valid Answer Rate (VAR)}, decoupling formatting reliability from domain knowledge and enabling fairer comparisons.

In controlled from-scratch ablations with 1.7B-parameter models, QVAC Genesis III yields large gains over Cosmopedia-v2 baselines and the publicly released Cosmo-1B model~\citep{cosmo1b} across ARC, GPQA, and MMLU STEM benchmarks, while the OL split achieves a near-perfect VAR of 99.45\% (Section~\ref{sec:experiments}). These results demonstrate that token-efficient, structurally rich synthetic data can substantially improve edge-scale model capabilities even under fixed compute budgets.

\section{Related work}
\label{sec:related}

\paragraph{Synthetic data for LLM pre-training.}
Scaling laws~\citep{hoffmann2022training} established that data quality and quantity jointly determine LLM performance, motivating efforts to generate high-quality synthetic corpora. Microsoft's Phi series was among the first to show that billions of synthetic tokens can train small models competitive with much larger systems trained on organic data~\citep{li2023phi15}, a finding especially relevant for edge-scale deployment where model size is constrained. Phi-4 extended this direction by incorporating multi-agent prompting, self-revision workflows, and instruction reversal, showing that synthetic-data quality, not merely volume, drives reasoning gains, and that student models can even surpass their teachers~\citep{abdin2024phi4}. HuggingFace's Cosmopedia provided the first large-scale \emph{open} replication of the Phi-style data recipe~\citep{benallal2024cosmopedia}, while concurrent work on web-corpus curation, including FineWeb~\citep{penedo2024fineweb} and Ultra-FineWeb~\citep{wang2025ultrafineweb}, developed quality classifiers for selecting high-value seed passages from large crawls. Despite these advances, existing open synthetic datasets remain limited in STEM coverage and are not optimized for the token efficiency demanded by edge-scale models, where every token in a tightly bounded pre-training budget must carry maximal learning signal.

\paragraph{Knowledge distillation and learning from model errors.}
Knowledge distillation~\citep{hinton2015distilling} transfers capabilities from a large teacher to a smaller student, typically by training on teacher-generated outputs. In the LLM setting, Orca~\citep{mukherjee2023orca} showed that progressive learning from detailed explanation traces produced by GPT-4 substantially improves small-model reasoning. More recently, \citet{burns2023weak} studied weak-to-strong generalization, finding that strong models can be elicited even from imperfect supervision. However, these approaches uniformly \emph{filter out} incorrect teacher or student outputs; the learning signal embedded in failures, misconceptions, reasoning errors, formatting ambiguities, is systematically discarded. Our Failure Analysis pipeline reverses this convention: rather than removing incorrect student responses, we use them as prompts for the teacher to generate targeted corrective explanations, converting each error into a focused teaching moment. Complementary to our approach, recent work on active synthetic data generation~\citep{kessler2025active} shows that iteratively curating data based on student performance outperforms static generation, though it focuses on fine-tuning rather than pre-training at scale.

\paragraph{Contrastive and option-level reasoning.}
Standard synthetic question-answering datasets retain only the correct solution trace, leaving the model to infer implicitly why alternative answers fail. Contrastive explanations, which explicitly articulate why distractors are incorrect, have been shown to improve generalization in smaller-scale educational settings~\citep{talmor2020teachingpretrained}. Our Option-Level Reasoning pipeline scales this principle to pre-training: for every correctly answered MCQ, the teacher generates exhaustive per-option justifications covering both the correct choice and each distractor, yielding richer supervision than single-trace solutions.

\paragraph{LLM-based evaluation of multiple-choice benchmarks.}
Multiple-choice evaluation of LLMs has traditionally relied on log-likelihood scoring over answer tokens. More recent frameworks such as OpenCompass~\citep{opencompass2023} support \emph{LLM-as-a-judge} protocols, where a separate model rates or grades open-ended outputs, following the paradigm introduced by \citet{zheng2023judging}. However, judging assesses \emph{quality} of a response, which is distinct from reliably \emph{extracting} a final committed answer from free-form text that may contain reasoning, hedging, or self-corrections. We extend OpenCompass with an \emph{LLM-as-a-parser} framework: rather than rating response quality, the parser is instructed solely to recover the model's final option choice, or to abstain when no unambiguous answer can be identified. This yields two decoupled metrics: standard accuracy and the Valid Answer Rate (VAR), which quantifies answer extractability independently of domain knowledge, a distinction invisible to both likelihood-based and judge-based evaluation.

\section{Methodology}
\label{sec:method}

\begin{figure}[t]
    \centering
    \includegraphics[width=0.85\linewidth]{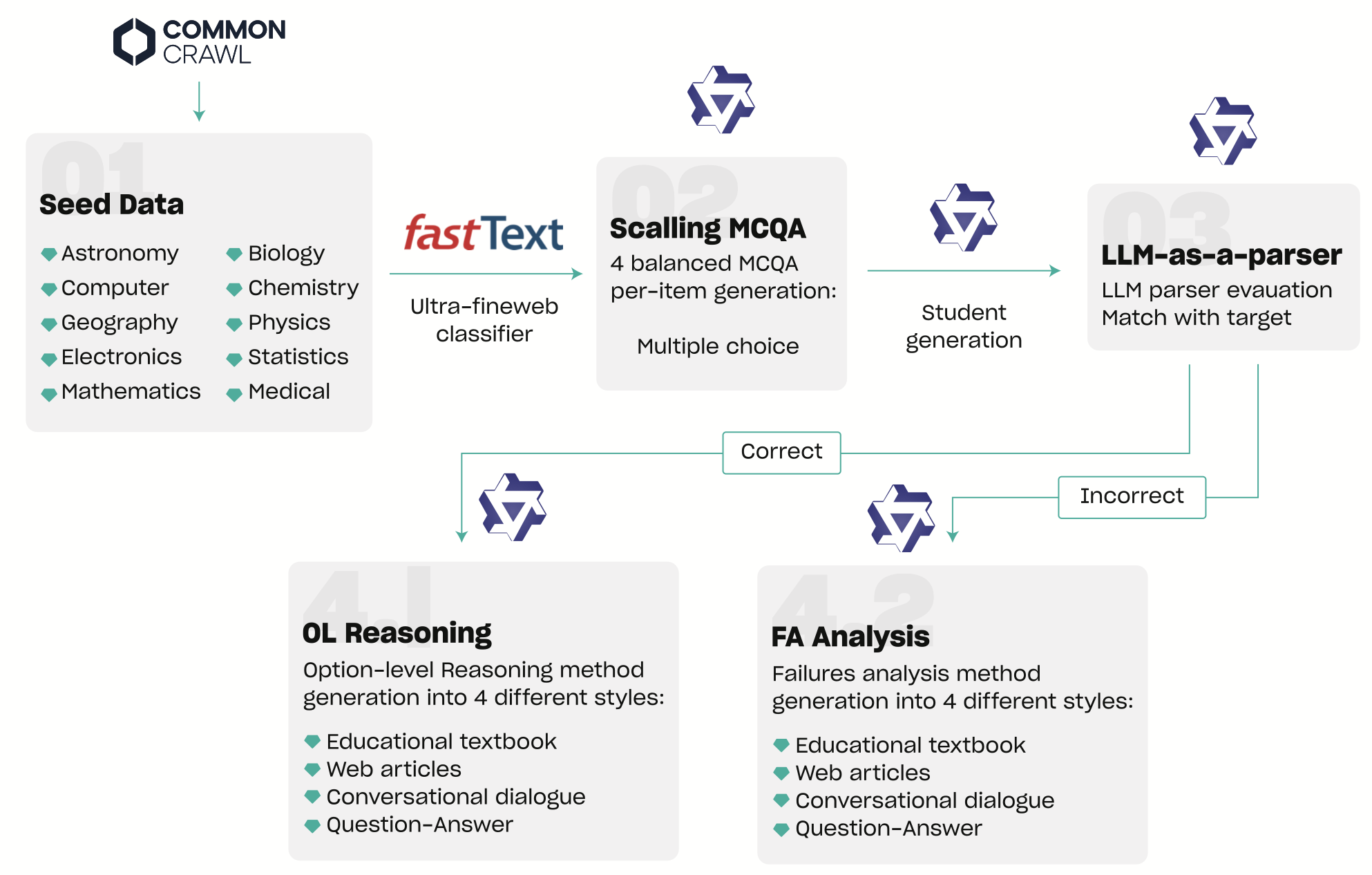}
    \caption{Overview of the dual-method synthetic data generation pipeline. Seed passages are quality-filtered, converted into MCQs, answered by a student model, and parsed via LLM-as-a-parser extraction. Correct answers are routed to Option-Level (OL) Reasoning; incorrect or non-extractable answers are routed to Failure Analysis (FA). Details in Section~\ref{sec:method}.}
    \label{fig:dual_pipeline}
\end{figure}

We construct QVAC Genesis III via a dual-method synthetic data pipeline that converts each generated multiple-choice question (MCQ) into pedagogically structured text (Figure~\ref{fig:dual_pipeline}). The pipeline consists of (i) seed acquisition and quality filtering, (ii) MCQ generation, (iii) model answering with answer extraction, and (iv) branching into either Failure Analysis (FA) or Option-Level (OL) Reasoning depending on whether the extracted answer matches the gold label.
All prompt templates used for MCQ generation, answering, answer extraction, and FA/OL rendering are provided in Appendix~\ref{app:prompts}.

\subsection{Seed acquisition and quality filtering}
\label{sec:seeds}

\paragraph{Seed acquisition.}
We source seed passages from FineFineWeb, a large-scale web corpus organized into coarse domains~\citep{zhang2024finefineweb}. We map each FineFineWeb domain to a set of curriculum-aligned generation subdomains used downstream (the full mapping is provided in Appendix~\ref{app:domain_map}); for example, FineFineWeb \texttt{biology} seeds are used to generate items for \texttt{college\_biology} and \texttt{high\_school\_biology}. We continue sampling candidates until we obtain approximately 500k high-quality seeds after filtering.

\paragraph{Quality filtering.}
Raw web text contains boilerplate, low-information content, and other noise that can degrade downstream synthetic data quality. We therefore apply the Ultra-FineWeb classifier~\citep{wang2025ultrafineweb}, a FastText-style binary classifier for quality discrimination, to each sampled passage and retain all instances classified as positive. The resulting positive set constitutes the seed pool used for MCQ generation.

\subsection{MCQ generation and format validation}
\label{sec:mcqgen}

\paragraph{MCQ generation.}
Given a filtered seed for a target domain and difficulty level, we generate self-contained MCQs using QwQ-32B~\citep{qwen_qwq32b} as the generator model. We selected QwQ-32B as generator and teacher model because, at 32B parameters, it can be deployed with high throughput while delivering competitive reasoning performance, making it a practical choice for large-scale synthetic data generation without requiring prohibitive infrastructure. Each generated item consists of a question, four mutually exclusive answer options (A--D), and a gold label indicating the correct choice. We enforce output constraints through a structured prompt template that requires exactly four options, a fixed CSV-like schema, and an even distribution of correct labels across A/B/C/D to prevent positional bias.

\paragraph{Post-processing and rejection.}
Despite prompt constraints, a fraction of generations violate formatting requirements (e.g., wrong number of fields or invalid label). We apply a lightweight Python post-processing script that validates each generated row and retains only correctly formatted instances. Approximately 15\% of raw generations are rejected by this validation step.

\subsection{Answering model and answer extraction}
\label{sec:answering}

\paragraph{Student generation.}
To obtain success/failure signals at scale, we require a student model representative of the edge-scale models that QVAC Genesis III targets. Our analysis indicates that 1--2B parameter models offer the best trade-off between accuracy and memory footprint / throughput for on-device deployment. We evaluated four pretrained base models in this range, Llama-3.2-1B, Gemma-3-1B, SmolLM2-1.7B, and Qwen3-1.7B-Base, and selected \textbf{Qwen3-1.7B-Base}~\citep{qwen3} as the data-generation student because it achieved the strongest baseline performance across our target STEM benchmarks (Appendix~\ref{app:additional_results}). Each generated MCQ is presented to this pretrained student, which produces a free-form response. This data-generation role is distinct from the controlled pretraining experiments in Section~\ref{sec:exp_setup}, whose checkpoints are initialized from random weights.

\paragraph{Answer extraction via model-based parsing.}
Student outputs frequently contain reasoning, hedging, or inconsistent formatting. Rather than relying on heuristic string parsing, we use CompassJudger-2-32B~\citep{zhang2025compassjudger}, a model fine-tuned specifically for judging and parsing tasks, as an \emph{extractor} (see also Figure~\ref{fig:llm_extractor}) that reads the student's full response and identifies the final option it selected. Despite its name, the model does not judge correctness: it only extracts the committed answer, which is subsequently compared with the gold label. The extractor returns a single option label (A--D) when a clear final answer can be identified, or abstains when the response is ambiguous or contains conflicting conclusions (corresponding to $\textsc{NoAns}$ and $\textsc{Multi}$ in the evaluation protocol of Section~\ref{sec:eval}). The same extraction procedure is applied uniformly across all domains and difficulty levels.

\paragraph{Branching policy.}
If the extracted answer matches the gold label, we treat the instance as a \emph{success} and route it to the Option-Level pipeline; otherwise (including cases where no unambiguous answer can be extracted), we treat it as a \emph{failure} and route it to Failure Analysis, since non-extractable responses provide useful signal about response clarity and reasoning failures.

\subsection{Failure Analysis (FA)}
\label{sec:fa}

Failure Analysis transforms incorrect or non-extractable student answers into corrective instructional content. For each failed instance, the teacher model (QwQ-32B~\citep{qwen_qwq32b}) receives the original question with all options, the student's full response, and the gold answer, and generates a pedagogical explanation that satisfies three requirements. First, \textbf{failure diagnosis}: the text explains why the student's response plausibly led to an incorrect or ambiguous choice and pinpoints the specific reasoning error, misconception, or formatting issue. Second, \textbf{correction}: the text provides a coherent solution that leads to the correct answer. Third, \textbf{self-containedness}: the full problem statement (question and all options) is embedded in the output so that the resulting document can be used as standalone training text without external context.
Outputs are generated in four complementary styles, educational textbook, web article, question-answer tutoring, and conversational dialogue, to maximize diversity in the resulting corpus.

\subsection{Option-Level reasoning (OL)}
\label{sec:ol}

Option-Level Reasoning converts correctly answered questions into contrastive, option-by-option explanations. While success cases are often retained only as a question--answer pair, OL produces richer supervision by explicitly analyzing \emph{every} answer choice. For each successful instance, the teacher model (QwQ-32B~\citep{qwen_qwq32b}) receives the question, all options, and the gold answer, and generates an explanation satisfying four requirements. First, \textbf{justification of the correct option}: the text provides a clear argument establishing why the correct choice holds, with step-by-step reasoning. Second, \textbf{refutation of distractors}: for every incorrect option, the text states a specific reason it fails (e.g., violated assumption, incorrect definition, wrong sign, or missing condition). Third, \textbf{deterministic coverage}: all options are addressed explicitly, leaving no choice unanalyzed, which reduces underspecification. Fourth, \textbf{self-containedness}: the full problem statement (question and all options) is included so the output serves as standalone training text.
As with FA, outputs are generated in four styles (textbook, web article, Q\&A, dialogue).
\begin{table}[t]
\caption{Overall corpus scale by split.}
\label{tab:qvac_totals}
\centering
\small
\begin{tabular}{lrr}
\toprule
\textbf{Split} & \textbf{Tokens} & \textbf{Documents} \\
\midrule
Option-Level (OL) & 108.67B & 92,538,646 \\
Failures (FA)     & 82.76B  & 67,107,907 \\
\midrule
Total             & 191.43B & 159,646,553 \\
\bottomrule
\end{tabular}
\end{table}

\paragraph{Corpus scale.}
The final corpus comprises two splits, OL (correct answers) and FA (incorrect/non-extractable answers), totalling \textbf{191.43B} tokens across \textbf{159.6M} documents (Table~\ref{tab:qvac_totals}; per-domain breakdowns in Appendix~\ref{app:corpus_detail}).

\subsection{Deduplication and decontamination}
\label{sec:dedup_decon}

\paragraph{Deduplication.}
We run MinHash near-deduplication over the full 159.6M-document corpus using 60-gram signatures. Out of 47,927 flagged duplicate pairs, only \textbf{1,729} unique documents are identified as near-duplicates ($<$0.002\% of the corpus), confirming that the synthetic generation pipeline produces highly diverse content with negligible redundancy.

\paragraph{Decontamination.}
We scan the corpus against all evaluation benchmarks using \textbf{Decon}~(AllenAI), a token-level $n$-gram overlap detector. A document is marked as contaminated when answer overlap $\geq$60\% with meaningful term matches or passage overlap $\geq$40\%. QVAC Genesis III contains only \textbf{17} verified contaminated documents across 159.6M (vs.\ 313 for Cosmopedia-v2 over 39.1M documents), all from the FA split where a small number of GSM8K and MMLU questions were echoed verbatim in model outputs during generation. The OL split has \textbf{zero} verified contamination. An initial 190 SQuAD matches were manually inspected and confirmed as false positives. Critically, test-set leakage is negligible: GSM8K test 0\%, MMLU test 0.005\% (3 of 56,168), ensuring that all reported benchmark scores are trustworthy. Full contamination statistics are provided in Table~\ref{tab:decon} (Appendix).

\section{Evaluation protocol}
\label{sec:eval}

\begin{figure}[t]
    \centering
    \includegraphics[width=\linewidth]{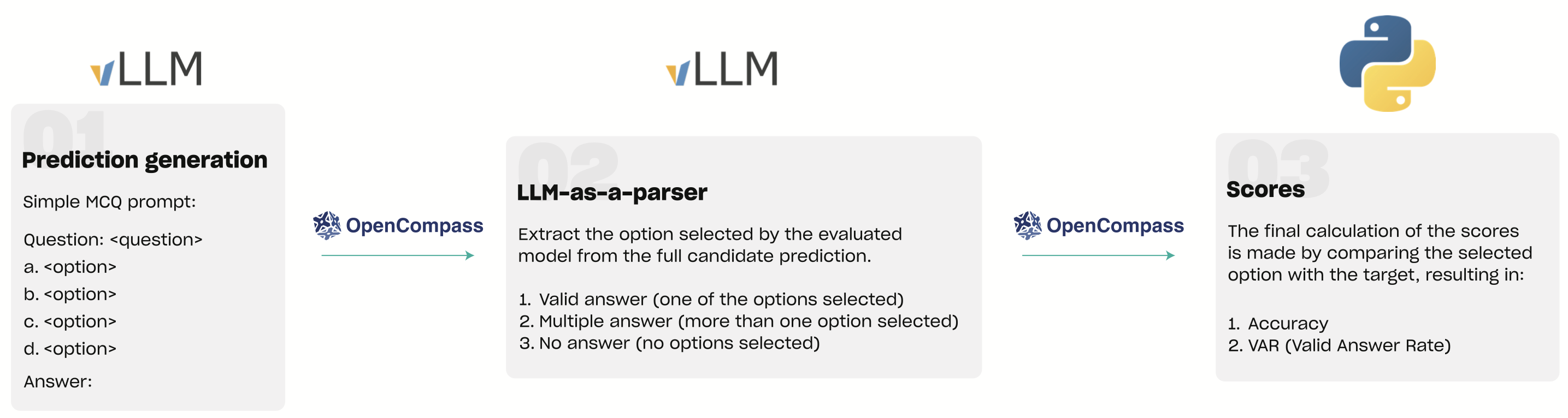}
    \caption{LLM-as-a-parser evaluation framework. The evaluated model generates a free-form prediction; a separate extractor LLM recovers the final committed option or abstains (\textsc{NoAns}/\textsc{Multi}), enabling decoupled reporting of Accuracy and VAR. Details in Section~\ref{sec:eval}.}
    \label{fig:llm_extractor}
\end{figure}

We build our evaluation on \textbf{OpenCompass}~\citep{opencompass2023}, modified to support a generation-first, answer-extraction workflow (Figure~\ref{fig:llm_extractor}). Rather than relying on log-likelihood scoring (which, while convenient, does not evaluate the model's realized generation and can be confounded by response-formatting issues, self-corrections, or ambiguous conclusions), we (i)~generate a full free-form answer, (ii)~\emph{parse} the text to extract a single final option, and (iii)~score correctness against the gold label. This protocol also exposes whether the model produces an unambiguous decision, which we quantify via the \emph{Valid Answer Rate (VAR)}.

\subsection{LLM-as-a-parser framework}
\label{sec:answer_extraction}

Given an MCQ item $x=(q,\{o_i\}_{i=1}^{m},y)$, the evaluated model produces a free-form response $r=A(x)$ (prompt in Appendix~\ref{app:mcq_answer}). A \emph{separate} extractor LLM then parses $r$ to recover the model's final committed choice; it is instructed \emph{not} to solve the question or assess correctness (prompt in Appendix~\ref{app:parser}). We use CompassJudger-2-32B~\citep{zhang2025compassjudger}, a model fine-tuned for judging and parsing tasks, as extractor under deterministic decoding (temperature~$0$). Correctness is determined only afterward by comparing the extracted option with the benchmark gold label. The extractor returns
\[
\hat{y} \;=\; J(r) \in \{1,\dots,m\} \cup \{\textsc{NoAns}, \textsc{Multi}\},
\]
where $\textsc{NoAns}$ indicates that no unambiguous final option can be extracted from $r$, and $\textsc{Multi}$ indicates that $r$ contains multiple conflicting final answers.

\subsection{Metrics: accuracy and extractability}
\label{sec:metrics}

Based on $\hat{y}$, we report two complementary metrics.

\paragraph{Valid Answer Rate (VAR).}
VAR measures the fraction of responses that contain an unambiguous, extractable final choice:
\begin{equation}
\mathrm{VAR}
=\frac{1}{N}\sum_{j=1}^{N}\mathbb{I}\!\left[\hat{y}_j \in \{1,\dots,m\}\right]
= 1 - \mathrm{NoAnsRate} - \mathrm{MultiRate}. \label{eq:var}
\end{equation}
Here $\mathrm{NoAnsRate}=\frac{1}{N}\sum_{j}\mathbb{I}[\hat{y}_j=\textsc{NoAns}]$ and $\mathrm{MultiRate}=\frac{1}{N}\sum_{j}\mathbb{I}[\hat{y}_j=\textsc{Multi}]$. VAR isolates answer extractability (decisiveness and formatting reliability) from domain knowledge. Appendix~\ref{app:additional_results} reports both failure rates separately and verifies that the main VAR conclusion is robust to two independent extractor models.

\paragraph{Accuracy.}
We report standard multiple-choice accuracy over all examples, treating non-extractable or conflicting outputs as incorrect:
$
\mathrm{Acc}
=\frac{1}{N}\sum_{j=1}^{N}\mathbb{I}\!\left[\hat{y}_j=y_j\right]. \label{eq:acc}
$
Together, VAR and accuracy separate ``can the model commit to a single option'' from ``is that option correct'', a distinction invisible to likelihood-based evaluation.

\section{Experiments}
\label{sec:experiments}

\subsection{Experimental setup}
\label{sec:exp_setup}

\paragraph{Experimental design.}
All models in our primary controlled comparisons use the Qwen3-1.7B architecture with identical hyperparameters and are trained from random initialization, rather than initialized from Qwen3-1.7B-Base, so that observed differences are attributable solely to data. We consider two settings:
\begin{enumerate}
    \item \textbf{Individual split ablation.} Separate models on the FA split (1\,ep, ${\approx}$82.8B tokens), the OL split (1\,ep, ${\approx}$108.7B tokens), and Cosmopedia-v2 (4\,ep, ${\approx}$109.7B tokens).
    \item \textbf{Combined corpus vs.\ baselines.} QVAC~Genesis III combined (1\,ep, 191.43B tokens) vs.\ (a) Cosmopedia-v2 (7\,ep, ${\approx}$192.4B tokens), providing a token-budget-matched comparison, and (b) \textbf{Cosmo-1B}~\citep{cosmo1b}, a 1.8B model pre-trained on 180B tokens of Cosmopedia augmented with code, mathematics, and instruction-following data.
\end{enumerate}
We additionally evaluate a same-stage checkpoint near one epoch of Cosmopedia-v2 and repeat the OL comparison with three non-Qwen backbones; full results are provided in Appendix~\ref{app:additional_results}.

\paragraph{Training framework and hyperparameters.}
All models are trained using \textbf{Megatron-Core}~\citep{shoeybi2019megatron} with \textbf{Megatron-Bridge} converting HuggingFace model definitions into Megatron-compatible format. Training uses Flash Attention~2, BF16 mixed precision, and sequence packing with end-of-document (\texttt{<EOD>}) attention-mask resets to prevent cross-document attention leakage. Each domain is loaded as a separate dataset with proportional-weights sampling so that each domain's contribution scales with its token count. All training is single-stage with no curriculum schedule; models are initialised from random weights. The full hardware and parallelism configuration is in Table~\ref{tab:parallelism} (Appendix). The full hyperparameter table including per-run wall-clock times and GPU-hours is provided in Table~\ref{tab:hparams} (Appendix).

\paragraph{Benchmarks.}
We evaluate on STEM-aligned benchmarks that match the training distribution:
\textbf{ARC-Easy/Challenge}~(ARC-E/ARC-C)~\citep{clark2018think} for science reasoning;
\textbf{GPQA Diamond}~\citep{rein2024gpqa} for graduate-level science QA; and
\textbf{MMLU STEM subsets}~\citep{hendrycks2021measuring}, covering the 19 STEM domains aligned with the QVAC~Genesis III curriculum (Appendix~\ref{app:domain_map}). We restrict MMLU to its STEM subdomains because the corpus does not cover non-STEM areas; evaluating on unrelated domains would conflate absence of training signal with data quality.
All benchmarks use the generation-first LLM-as-a-parser protocol (Section~\ref{sec:eval}), reporting both accuracy and VAR.

\subsection{Individual split ablation (from scratch)}
\label{sec:main_results}

We isolate each pipeline component by training separate models on the FA split (1~epoch, 82.76B tokens), the OL split (1~epoch, 108.67B tokens), and Cosmopedia-v2 (4~epochs, ${\approx}$109.7B tokens). Token budgets are closely matched for OL and Cosmopedia-v2; the FA run is ${\sim}$25\% smaller by design. Differences therefore primarily reflect data quality rather than scale.

Table~\ref{tab:main_results}(a) reports results. The OL split delivers the largest gains: \textbf{+26.81} on ARC-E, \textbf{+15.25} on ARC-C, and \textbf{+10.11} on GPQA Diamond over Cosmopedia-v2. FA also outperforms Cosmopedia-v2 on all three benchmarks (+9.18 ARC-E, +2.03 ARC-C, +3.54 GPQA), indicating that learning from model errors provides meaningful signal even in isolation. On MMLU STEM, OL achieves \textbf{30.26\%} accuracy (${\approx}$+10 over Cosmopedia-v2) and a VAR of \textbf{99.45\%}, demonstrating that the contrastive option-level format improves both domain knowledge and the model's ability to produce unambiguous outputs.

The additional controls support the same conclusion (Appendix~\ref{app:additional_results}). At the closest same-stage checkpoint (${\approx}$25B Cosmopedia-v2 tokens), FA and OL improve MMLU by 13.65 and 20.80 points, respectively, over Cosmopedia-v2; OL also improves ARC-C/ARC-E by 15.93/16.94 points. Across Llama-3.2-1B, SmolLM2-1.7B, and Gemma3-1B, QVAC-OL outperforms the corresponding Cosmopedia-v2 run on every reported benchmark, indicating that the gain is not specific to the Qwen architecture.

\begin{table}[t]
    \caption{Main results. \textit{Panel~(a):} individual split ablation, all models 1.7B parameters, trained from scratch (FA: 1\,ep / 82.76B tokens; OL: 1\,ep / 108.67B tokens; Cosmopedia-v2: 4\,ep / ${\approx}$109.7B tokens). \textit{Panel~(b):} combined corpus vs.\ baselines, QVAC Genesis III Combined (1\,ep / 191.43B tokens) vs.\ token-matched Cosmopedia-v2 (7\,ep / ${\approx}$192.4B tokens) and Cosmo-1B. All values are accuracy (\%) or VAR (\%).}
    \label{tab:main_results}
    \centering
    \begin{small}
    \begin{tabular}{lccccc}
    \toprule
    & \multicolumn{3}{c}{\textbf{Science reasoning}} & \multicolumn{2}{c}{\textbf{MMLU STEM (19 dom.)}} \\
    \cmidrule(lr){2-4} \cmidrule(lr){5-6}
    \textbf{Model / Training data} & \textbf{ARC-E} & \textbf{ARC-C} & \textbf{GPQA} & \textbf{Acc} & \textbf{VAR} \\
    \midrule
    \multicolumn{6}{l}{\textit{(a) Individual split ablation}} \\
    \addlinespace[2pt]
    Cosmopedia-v2 (4\,ep)       & 20.63 & 21.02 & 17.67 & 20.39 & 72.71 \\
    QVAC Genesis III -- FA           & 29.81 & 23.05 & 21.21 & 23.29 & 78.14 \\
    QVAC Genesis III -- OL           & \textbf{47.44} & \textbf{36.27} & \textbf{27.78} & \textbf{30.26} & \textbf{99.45} \\
    \cmidrule(lr){1-6}
    $\Delta$ (OL vs.\ Cosmo-v2 4\,ep) & +26.81 & +15.25 & +10.11 & +9.87 & +26.74 \\
    \midrule\midrule
    \multicolumn{6}{l}{\textit{(b) Combined corpus vs.\ baselines}} \\
    \addlinespace[2pt]
    Cosmopedia-v2 (7\,ep) & 23.28 & 21.36 & 20.20 & 15.16 & 62.48 \\
    Cosmo-1B                       & 28.04 & 23.73 & 19.70 & 25.42 & \textbf{92.51} \\
    QVAC Genesis III Combined (FA+OL)  & \textbf{51.85} & \textbf{42.71} & \textbf{22.72} & \textbf{30.19} & 92.06 \\
    \cmidrule(lr){1-6}
    $\Delta$ (vs.\ Cosmo-v2 7\,ep)  & +28.57 & +21.35 & +2.52  & +15.03 & +29.58 \\
    $\Delta$ (vs.\ Cosmo-1B)        & +23.81 & +18.98 & +3.02  & +4.77  & $-$0.45 \\
    \bottomrule
    \end{tabular}
    \end{small}
\end{table}

\subsection{Combined corpus vs.\ baselines}
\label{sec:cosmo1b}

We now evaluate the full QVAC Genesis III corpus (FA+OL combined, 1 epoch, 191.43B tokens) against the two baselines described in Section~\ref{sec:exp_setup}: a token-budget-matched Cosmopedia-v2 run (7 epochs, ${\approx}$192.4B tokens) and the publicly released Cosmo-1B~\citep{cosmo1b}.

Table~\ref{tab:main_results}(b) reports results (per-domain breakdown in Table~\ref{tab:mmlu_all_detail}). Against the token-matched Cosmopedia-v2 baseline, QVAC Genesis III Combined achieves \textbf{+28.57} on ARC-E, \textbf{+21.35} on ARC-C, and \textbf{+15.03} on MMLU STEM, with a VAR improvement of \textbf{+29.58} points (92.06\% vs.\ 62.48\%). Notably, the 7-epoch Cosmopedia-v2 model shows lower accuracy and degraded VAR compared to its 4-epoch counterpart, suggesting that repeated passes over the same ${\approx}$27.5B unique tokens cause overfitting that harms both knowledge and answer quality.

Against Cosmo-1B, QVAC Genesis III Combined outperforms by \textbf{+23.81} on ARC-E, \textbf{+18.98} on ARC-C, \textbf{+3.02} on GPQA Diamond, and \textbf{+4.77} on MMLU STEM, while maintaining comparable VAR (92.06\% vs.\ 92.51\%). Cosmo-1B is not a controlled Cosmopedia-v2-only baseline: it benefits from additional code, mathematics, instruction-following, and chat-formatted data, whereas our QVAC checkpoints are trained from scratch using pretraining data only.

\subsection{Analysis and per-domain breakdown}
\label{sec:mmlu_detail}

Table~\ref{tab:mmlu_all_detail} provides the full per-domain breakdown. Three patterns stand out. First, OL is the stronger standalone split overall and outperforms FA in 18 of 19 domains, plausibly because it analyzes every answer option and is generated from easier, correctly answered questions. FA instead targets harder student failures and provides corrective signal; it is strongest relative to OL in Professional~Medicine (28.7\% vs.\ 19.9\%). These signals can still complement one another: on High School Geography, Combined reaches 35.4\%, versus 31.8\% for OL and 25.8\% for FA. Second, when the splits are combined, QVAC~Genesis III outperforms Cosmo-1B on 12 of 19 domains, with the largest margins in biological and natural sciences (College~Bio +18.7, HS~Bio +16.1); Cosmo-1B retains an edge primarily in formally structured domains (Econometrics, CS, Physics) where its code and math augmentation provides additional signal. Third, VAR patterns reveal that data structure directly affects output quality: OL achieves 100\% VAR on 15 of 19 domains, while Cosmopedia-v2 VAR degrades from 72.71\% (4~epochs) to 62.48\% (7~epochs), indicating that data diversity, not repetition, is key to maintaining answer extractability.

\begin{table}[!t]
    \caption{Per-domain MMLU STEM results across all settings. \textit{Left:} individual split ablation (Cosmopedia-v2 4\,ep; FA 1\,ep; OL 1\,ep). \textit{Right:} combined corpus vs.\ baselines (Cosmopedia-v2 7\,ep; Cosmo-1B; QVAC Combined). All values in \%.}
    \label{tab:mmlu_all_detail}
    \centering
    \setlength{\tabcolsep}{3pt}
    \renewcommand{\arraystretch}{1.05}
    \adjustbox{max width=\linewidth}{%
    \begin{tabular}{l cc cc cc @{\hskip 4pt}!{\vrule width 0.6pt}@{\hskip 4pt} cc cc cc}
    \toprule
    & \multicolumn{6}{c}{\textbf{Individual split ablation}} & \multicolumn{6}{c}{\textbf{Combined vs.\ baselines}} \\
    \cmidrule(lr){2-7} \cmidrule(l){8-13}
    & \multicolumn{2}{c}{\textbf{Cosmo-v2\,(4\,ep)}} & \multicolumn{2}{c}{\textbf{QVAC--FA}} & \multicolumn{2}{c}{\textbf{QVAC--OL}}
    & \multicolumn{2}{c}{\textbf{Cosmo-v2\,(7\,ep)}} & \multicolumn{2}{c}{\textbf{Cosmo-1B}} & \multicolumn{2}{c}{\textbf{QVAC\,Comb.}} \\
    \cmidrule(lr){2-3} \cmidrule(lr){4-5} \cmidrule(lr){6-7}
    \cmidrule(lr){8-9} \cmidrule(lr){10-11} \cmidrule(l){12-13}
    \textbf{Domain} & Acc & VAR & Acc & VAR & Acc & VAR & Acc & VAR & Acc & VAR & Acc & VAR \\
    \midrule
    Astronomy            & 25.7 & 70.4 & 21.7 & 73.7 & \textbf{34.9} & \textbf{100.0} & 17.1 & 78.3 & 24.3 & 88.2 & \textbf{39.5} & \textbf{93.4} \\
    Elec.\ Eng.          & 22.1 & 80.7 & 28.3 & 77.2 & \textbf{35.9} & \textbf{100.0} & 21.4 & 71.7 & \textbf{31.0} & \textbf{95.2} & 27.6 & 93.8 \\
    HS Geography         & 20.2 & 80.3 & 25.8 & 73.7 & \textbf{31.8} & \textbf{100.0} & 21.2 & 77.3 & 27.8 & 91.9 & \textbf{35.4} & \textbf{93.9} \\
    College Bio.         & 21.5 & 72.2 & 23.6 & 66.7 & \textbf{30.6} & \textbf{100.0} & 18.8 & 63.9 & 18.1 & 88.9 & \textbf{36.8} & \textbf{93.8} \\
    HS Biology           & 21.6 & 71.6 & 17.7 & 73.2 & \textbf{35.8} & \textbf{100.0} & 21.6 & 69.4 & 23.6 & 91.3 & \textbf{39.7} & \textbf{94.5} \\
    College Med.         & 17.3 & 65.3 & 23.1 & 78.0 & \textbf{34.7} & \textbf{100.0} & 15.6 & 68.2 & 24.9 & 91.9 & \textbf{31.2} & \textbf{94.2} \\
    Prof.\ Med.          & 27.2 & 79.8 & \textbf{28.7} & 94.5 & 19.9 & \textbf{100.0} & 14.7 & 71.0 & 17.3 & 91.2 & \textbf{25.0} & \textbf{96.0} \\
    College Maths        & 22.0 & 56.0 & 17.0 & 71.0 & \textbf{27.0} & \textbf{94.0}  & 12.0 & 47.0 & \textbf{25.0} & \textbf{99.0} & 24.0 & 84.0 \\
    HS Mathematics       & 16.7 & 74.8 & 25.6 & 90.4 & \textbf{28.2} & \textbf{100.0} & 12.2 & 50.7 & 24.8 & \textbf{92.6} & \textbf{25.2} & 85.9 \\
    College Physics      & 12.8 & 74.5 & 23.5 & 90.2 & \textbf{24.5} & \textbf{100.0} & 15.7 & 63.7 & \textbf{30.4} & 94.1 & 28.4 & \textbf{96.1} \\
    HS Physics           & 17.2 & 74.2 & 21.2 & 70.9 & \textbf{29.8} & \textbf{100.0} & 19.2 & 63.6 & \textbf{30.5} & \textbf{96.0} & 28.5 & 95.4 \\
    Concept.\ Physics    & 22.6 & 76.6 & 26.8 & 82.6 & \textbf{33.6} & \textbf{100.0} & 17.4 & 66.0 & 28.1 & 93.6 & \textbf{34.0} & \textbf{94.9} \\
    College Chem.        & \textbf{29.0} & 76.0 & 25.0 & 77.0 & 26.0 & \textbf{100.0} & 5.0  & 53.0 & 18.0 & 88.0 & \textbf{31.0} & \textbf{94.0} \\
    HS Chemistry         & 12.3 & 65.5 & 15.8 & 74.4 & \textbf{33.0} & \textbf{99.5}  & 15.8 & 59.6 & 27.1 & 92.6 & \textbf{29.6} & \textbf{96.1} \\
    College CS           & 10.0 & 56.0 & 23.0 & 80.0 & \textbf{31.0} & \textbf{99.0}  & 10.0 & 54.0 & \textbf{30.0} & \textbf{96.0} & 25.0 & 90.0 \\
    HS Comp.\ Sci.       & 16.0 & 61.0 & 25.0 & 79.0 & \textbf{36.0} & \textbf{100.0} & 13.0 & 63.0 & \textbf{32.0} & \textbf{92.0} & 30.0 & 83.0 \\
    Machine Learning     & 18.8 & 66.1 & 24.1 & 71.4 & \textbf{30.4} & \textbf{99.1}  & 14.3 & 53.6 & 18.8 & \textbf{93.8} & \textbf{31.3} & 80.4 \\
    HS Statistics        & 16.2 & 61.6 & 13.9 & 71.3 & \textbf{24.1} & \textbf{100.0} & 12.5 & 61.6 & 20.8 & 91.2 & \textbf{27.8} & \textbf{97.7} \\
    Econometrics         & 20.2 & 56.1 & 17.5 & 65.8 & \textbf{24.6} & \textbf{100.0} & 10.5 & 51.8 & \textbf{30.7} & 90.4 & 23.7 & \textbf{92.1} \\
    \midrule
    \textbf{Average}     & 20.4 & 72.7 & 23.3 & 78.1 & \textbf{30.3} & \textbf{99.5}  & 15.2 & 62.5 & 25.4 & \textbf{92.5} & \textbf{30.2} & 92.1 \\
    \bottomrule
    \end{tabular}%
    }
\end{table}

\section{Conclusion}
\label{sec:conclusion}

We presented QVAC Genesis III, a 191.43B-token STEM-focused synthetic corpus that converts both model failures and successes into structured educational content through targeted teacher distillation, designed to maximize token efficiency for edge-scale language models. Our key contributions are: (i)~\textbf{QVAC Genesis III}, a multi-domain corpus spanning 19 curriculum-aligned domains and three difficulty levels; (ii)~\textbf{Failure Analysis}, a teacher-distilled procedure that transforms incorrect student responses into targeted corrective explanations; (iii)~\textbf{Option-Level Reasoning}, a contrastive generation strategy that produces exhaustive per-option justifications; and (iv)~an \textbf{LLM-as-a-parser evaluation framework} that decouples answer validity from domain knowledge via the Valid Answer Rate metric.
To validate the effectiveness of our contributions, we conducted experiments with 1.7B-parameter models and demonstrated that our QVAC Genesis III data substantially outperforms both Cosmopedia-v2 baselines and the Cosmo-1B model across all evaluated STEM benchmarks, with the OL split achieving near-perfect answer extractability (99.45\% VAR). Human domain-expert assessment of factual correctness, pedagogical usefulness, and hallucinations in the generated explanations remains important future work. To support future research, we will release the QVAC Genesis III corpus under CC-BY-NC-4.0 and the FA, OL, and Combined pretrained models under Apache-2.0, offering a comprehensive foundation for building efficient education and STEM small models targeting edge and on-device deployment.

\small
\bibliographystyle{unsrtnat}
\bibliography{bibliography}

@inproceedings{hoffmann2022training,
  title={An empirical analysis of compute-optimal large language model training},
  author={Hoffmann, Jordan and Borgeaud, Sebastian and Mensch, Arthur and Buchatskaya, Elena and Cai, Trevor and Rutherford, Eliza and de Las Casas, Diego and Hendricks, Lisa Anne and Welbl, Johannes and Clark, Aidan and others},
  booktitle={Advances in Neural Information Processing Systems},
  volume={35},
  pages={30016--30030},
  year={2022}
}

@article{li2023phi15,
  title={Textbooks Are All You Need {II}: phi-1.5 Technical Report},
  author={Li, Yuanzhi and Bubeck, S{\'e}bastien and Eldan, Ronen and Del Giorno, Allie and Gunasekar, Suriya and Lee, Yin Tat},
  journal={arXiv preprint arXiv:2309.05463},
  year={2023}
}

@article{abdin2024phi4,
  title={Phi-4 Technical Report},
  author={Abdin, Marah and Aneja, Jyoti and Behl, Harkirat and Bubeck, S{\'e}bastien and Eldan, Ronen and Gunasekar, Suriya and Harrison, Michael and Hewett, Russell J. and Javaheripi, Mojan and Kauffmann, Piero and Lee, James R. and Lee, Yin Tat and Li, Yuanzhi and Liu, Weishung and Mendes, Caio C. T. and Nguyen, Anh and Price, Eric and de Rosa, Gustavo and Saarikivi, Olli and Salim, Adil and Shah, Shital and Wang, Xin and Ward, Rachel and Wu, Yue and Yu, Dingli and Zhang, Cyril and Zhang, Yi},
  journal={arXiv preprint arXiv:2412.08905},
  year={2024}
}

@misc{benallal2024cosmopedia,
  title={Cosmopedia: how to create large-scale synthetic data for pre-training Large Language Models},
  author={Ben Allal, Loubna and Lozhkov, Anton and van Strien, Daniel},
  howpublished={\url{https://huggingface.co/blog/cosmopedia}},
  month={March},
  year={2024}
}

@inproceedings{penedo2024fineweb,
  title={The {FineWeb} Datasets: Decanting the Web for the Finest Text Data at Scale},
  author={Penedo, Guilherme and Kydl{\'i}{\v{c}}ek, Hynek and Ben Allal, Loubna and Lozhkov, Anton and Mitchell, Margaret and Raffel, Colin and von Werra, Leandro and Wolf, Thomas},
  booktitle={Advances in Neural Information Processing Systems},
  volume={37},
  year={2024}
}

@article{wang2025ultrafineweb,
  title={{Ultra-FineWeb}: Efficient Data Filtering and Verification for High-Quality {LLM} Training Data},
  author={Wang, Yudong and Fu, Zixuan and Cai, Jie and Tang, Peijun and Lyu, Hongya and Fang, Yewei and Zheng, Zhi and Zhou, Jie and Zeng, Guoyang and Xiao, Chaojun and Han, Xu and Liu, Zhiyuan},
  journal={arXiv preprint arXiv:2505.05427},
  year={2025}
}

@misc{zhang2024finefineweb,
  title={{FineFineWeb}: A Comprehensive Study on Fine-grained Domain Web Corpus},
  author={{M-A-P} and Zhang, Ge and Du, Xinrun and Yu, Zhimiao and Wang, Zili and Wang, Zekun and others},
  howpublished={\url{https://huggingface.co/datasets/m-a-p/FineFineWeb}},
  year={2024}
}

@article{hinton2015distilling,
  title={Distilling the Knowledge in a Neural Network},
  author={Hinton, Geoffrey and Vinyals, Oriol and Dean, Jeff},
  journal={arXiv preprint arXiv:1503.02531},
  year={2015}
}

@article{mukherjee2023orca,
  title={Orca: Progressive Learning from Complex Explanation Traces of {GPT}-4},
  author={Mukherjee, Subhabrata and Mitra, Arindam and Jawahar, Ganesh and Agarwal, Sahaj and Palangi, Hamid and Awadallah, Ahmed},
  journal={arXiv preprint arXiv:2306.02707},
  year={2023}
}

@inproceedings{burns2023weak,
  title={Weak-to-Strong Generalization: Eliciting Strong Capabilities With Weak Supervision},
  author={Burns, Collin and Izmailov, Pavel and Kirchner, Jan Hendrik and Baker, Bowen and Gao, Leo and Aschenbrenner, Leopold and Chen, Yining and Ecoffet, Adrien and Joglekar, Manas and Leike, Jan and Sutskever, Ilya and Wu, Jeffrey},
  booktitle={Proceedings of the 41st International Conference on Machine Learning (ICML)},
  volume={235},
  series={Proceedings of Machine Learning Research},
  pages={4971--5012},
  publisher={PMLR},
  year={2024}
}

@inproceedings{talmor2020teachingpretrained,
  title={{Leap-Of-Thought}: Teaching Pre-Trained Models to Systematically Reason Over Implicit Knowledge},
  author={Talmor, Alon and Tafjord, Oyvind and Clark, Peter and Goldberg, Yoav and Berant, Jonathan},
  booktitle = {Neural Information Processing Systems},
  volume={33},
  pages={20227--20237},
  year={2020},
  journal = {Neural Information Processing Systems},
}

@article{kessler2025active,
  title={Towards Active Synthetic Data Generation for Finetuning Language Models},
  author={Kessler, Samuel and Xia, Menglin and Madrigal Diaz, Daniel and Han, Dongge and Hashemi, Helia and Rajmohan, Saravan and R{\"u}hle, Victor and Ash, Jordan T.},
  journal={arXiv preprint arXiv:2512.00884},
  year={2025}
}

@misc{opencompass2023,
  title={{OpenCompass}: A Universal Evaluation Platform for Foundation Models},
  author={{OpenCompass Contributors}},
  howpublished={\url{https://github.com/open-compass/opencompass}},
  year={2023}
}

@inproceedings{zheng2023judging,
  title={Judging {LLM}-as-a-Judge with {MT}-Bench and Chatbot Arena},
  author={Zheng, Lianmin and Chiang, Wei-Lin and Sheng, Ying and Zhuang, Siyuan and Wu, Zhanghao and Zhuang, Yonghao and Lin, Zi and Li, Zhuohan and Li, Dacheng and Xing, Eric P. and others},
  booktitle={Advances in Neural Information Processing Systems},
  volume={36},
  year={2023}
}

@article{zhang2025compassjudger,
  title={CompassJudger-2: Towards Generalist Judge Model via Verifiable Rewards},
  author={Zhang, Taolin and Cao, Maosong and Lam, Alexander and Zhang, Songyang and Chen, Kai},
  journal={arXiv preprint arXiv:2507.09104},
  year={2025}
}

@misc{qwen_qwq32b,
  title={{QwQ-32B}: Embracing the Power of Reinforcement Learning},
  author={{Qwen Team}},
  howpublished={\url{https://qwenlm.github.io/blog/qwq-32b/}},
  year={2025}
}

@article{qwen3,
  title={Qwen3 Technical Report},
  author={Yang, An and others},
  journal={arXiv preprint arXiv:2505.09388},
  year={2025}
}

@misc{cosmo1b,
  title={Cosmo-1B},
  author={{HuggingFace SmolModels Team}},
  howpublished={\url{https://huggingface.co/HuggingFaceTB/cosmo-1b}},
  year={2024}
}

@article{clark2018think,
  title={Think you have Solved Question Answering? Try {ARC}, the {AI2} Reasoning Challenge},
  author={Clark, Peter and Cowhey, Isaac and Etzioni, Oren and Khot, Tushar and Sabharwal, Ashish and Schoenick, Carissa and Tafjord, Oyvind},
  journal={arXiv preprint arXiv:1803.05457},
  year={2018}
}

@inproceedings{rein2024gpqa,
  title={{GPQA}: A Graduate-Level {Google}-Proof {Q\&A} Benchmark},
  author={Rein, David and Hou, Betty Li and Stickland, Asa Cooper and Petty, Jackson and Pang, Richard Yuanzhe and Dirani, Julien and Michael, Julian and Bowman, Samuel R.},
  booktitle={Proceedings of the First Conference on Language Modeling},
  year={2024}
}

@inproceedings{hendrycks2021measuring,
  title={Measuring Massive Multitask Language Understanding},
  author={Hendrycks, Dan and Burns, Collin and Basart, Steven and Zou, Andy and Mazeika, Mantas and Song, Dawn and Steinhardt, Jacob},
  booktitle={International Conference on Learning Representations},
  year={2021}
}

@article{shoeybi2019megatron,
  title={Megatron-{LM}: Training Multi-Billion Parameter Language Models Using Model Parallelism},
  author={Shoeybi, Mohammad and Patwary, Mostofa and Puri, Raul and LeGresley, Patrick and Casper, Jared and Catanzaro, Bryan},
  journal={arXiv preprint arXiv:1909.08053},
  year={2019}
}
\normalsize

\clearpage
\appendix
\let\oldsection\section
\renewcommand{\section}{\FloatBarrier\oldsection}

\section{Domain mapping}
\label{app:domain_map}

Table~\ref{tab:domain_map} shows the mapping from FineFineWeb source domains to the curriculum-aligned generation subdomains used in QVAC Genesis III.

\begin{table}[htbp]
    \caption{Mapping from FineFineWeb domains to curriculum-aligned generation subdomains used in QVAC Genesis III.}
    \label{tab:domain_map}
    \centering
    \small
    \renewcommand{\arraystretch}{1.08}
    \begin{tabular}{@{}ll@{}}
        \toprule
        \textbf{FineFineWeb domain} & \textbf{QVAC Genesis III domain} \\
        \midrule
        \texttt{astronomy} &
        \texttt{astronomy} \\

        \texttt{biology} &
        \makecell[l]{\texttt{college\_biology}\\\texttt{high\_school\_biology}} \\

        \texttt{chemistry} &
        \makecell[l]{\texttt{college\_chemistry}\\\texttt{high\_school\_chemistry}} \\

        \texttt{computer science} &
        \makecell[l]{\texttt{college\_computer\_science}\\\texttt{high\_school\_computer\_science}\\\texttt{machine\_learning}} \\

        \texttt{geography} &
        \texttt{high\_school\_geography} \\

        \texttt{physics} &
        \makecell[l]{\texttt{college\_physics}\\\texttt{high\_school\_physics}\\\texttt{conceptual\_physics}} \\

        \texttt{statistics} &
        \makecell[l]{\texttt{high\_school\_statistics}\\\texttt{econometrics}} \\

        \texttt{electronic science} &
        \texttt{electrical\_engineering} \\

        \texttt{medical} &
        \makecell[l]{\texttt{college\_medicine}\\\texttt{professional\_medicine}} \\

        \texttt{mathematics} &
        \makecell[l]{\texttt{college\_maths}\\\texttt{high\_school\_maths}} \\
        \bottomrule
    \end{tabular}
\end{table}

\section{Per-domain corpus scale}
\label{app:corpus_detail}

Table~\ref{tab:corpus_per_domain} reports the token and document counts for each domain in both the OL and FA splits.

\begin{table}[htbp]
    \caption{Per-domain corpus scale for Option-Level (OL) and Failure Analysis (FA) splits.}
    \label{tab:corpus_per_domain}
    \centering
    \small
    \setlength{\tabcolsep}{5pt}
    \renewcommand{\arraystretch}{1.02}
    \begin{tabular}{l rr rr}
    \toprule
    & \multicolumn{2}{c}{\textbf{Option-Level (OL)}} & \multicolumn{2}{c}{\textbf{Failure Analysis (FA)}} \\
    \cmidrule(lr){2-3} \cmidrule(lr){4-5}
    \textbf{Domain} & Tokens & Docs & Tokens & Docs \\
    \midrule
    College Medicine       & 13.37B & 11.6M & 6.22B & 5.2M \\
    HS Biology             & 11.12B & 9.9M  & 4.51B & 3.8M \\
    College Biology        & 8.69B  & 7.6M  & 3.93B & 3.3M \\
    College Maths          & 7.44B  & 5.8M  & 7.58B & 5.7M \\
    Machine Learning       & 5.50B  & 4.6M  & 3.82B & 3.1M \\
    HS Statistics          & 5.31B  & 4.4M  & 4.16B & 3.3M \\
    HS Mathematics         & 5.29B  & 4.2M  & 4.04B & 3.2M \\
    HS Chemistry           & 5.18B  & 4.5M  & 4.02B & 3.3M \\
    College Chemistry      & 4.93B  & 4.2M  & 4.82B & 3.9M \\
    HS Geography            & 4.76B  & 4.3M  & 4.57B & 4.0M \\
    Econometrics           & 4.73B  & 3.8M  & 4.43B & 3.4M \\
    Astronomy              & 4.62B  & 4.0M  & 4.47B & 3.7M \\
    College Physics        & 4.49B  & 3.6M  & 5.16B & 4.0M \\
    College CS             & 4.45B  & 3.8M  & 4.68B & 3.9M \\
    HS Comp.\ Science      & 4.40B  & 3.9M  & 4.01B & 3.4M \\
    Elec.\ Engineering     & 4.37B  & 3.7M  & 4.74B & 3.9M \\
    Prof.\ Medicine        & 3.89B  & 3.3M  & 1.87B & 1.5M \\
    HS Physics             & 3.25B  & 2.7M  & 2.85B & 2.2M \\
    Conceptual Physics     & 2.90B  & 2.5M  & 2.88B & 2.3M \\
    \midrule
    \textbf{Total}         & \textbf{108.67B} & \textbf{92.5M} & \textbf{82.76B} & \textbf{67.1M} \\
    \bottomrule
    \end{tabular}
\end{table}

\section{Training configuration and contamination details}
\label{app:training_details}

Training runs on 64 NVIDIA H100 (80\,GB) GPUs across 8 nodes connected via InfiniBand with GPU Direct RDMA, using Tensor Parallelism (TP\,=\,2) over NVLink, Data Parallelism (DP\,=\,32) across nodes, and no Pipeline Parallelism (PP\,=\,1).

All runs use AdamW ($\beta_1{=}0.9$, $\beta_2{=}0.95$, $\varepsilon{=}10^{-5}$) with cosine learning-rate decay from $2{\times}10^{-4}$ to $2{\times}10^{-5}$, 10\% warmup, a global batch size of 2,048 sequences (${\approx}$8.4M tokens/step), and BF16 precision. Total training time ranges from ${\sim}$13.5\,h (FA) to ${\sim}$31.5\,h (Combined) on 64$\times$H100, with the matched-budget Cosmopedia-v2 baseline taking ${\sim}$29.8\,h. 

\begin{table}[htbp]
\caption{Hardware and parallelism configuration.}
\label{tab:parallelism}
\centering
\small
\begin{tabular}{ll}
\toprule
\textbf{Setting} & \textbf{Value} \\
\midrule
Hardware                  & 64\,$\times$\,H100 (80\,GB), 8 nodes \\
Tensor Parallelism (TP)   & 2 \\
Pipeline Parallelism (PP) & 1 \\
Data Parallelism (DP)     & 32 \\
Interconnect              & InfiniBand + GPU Direct RDMA \\
\bottomrule
\end{tabular}
\end{table}

\begin{table}[h]
\caption{Training hyperparameters (identical across all runs).}
\label{tab:hparams}
\centering
\small
\begin{tabular}{ll}
\toprule
\textbf{Hyperparameter} & \textbf{Value} \\
\midrule
Architecture             & Qwen3-1.7B (random init) \\
Sequence length          & 4,096 tokens \\
Optimizer                & AdamW ($\beta_1{=}0.9$, $\beta_2{=}0.95$, $\varepsilon{=}10^{-5}$) \\
Global batch size        & 2,048 sequences (${\approx}8.4$M tokens/step) \\
Micro batch size         & 4 per GPU \\
Gradient accumulation    & 16 steps \\
Learning rate            & $2{\times}10^{-4} \to 2{\times}10^{-5}$ (cosine decay) \\
Warmup                   & 10\% of total steps \\
Weight decay             & 0.01 \\
Gradient clipping        & 1.0 \\
Precision                & BF16 \\
\multicolumn{2}{l}{\textit{Training duration}} \\
\quad FA (Failures only)             & 1 epoch (${\approx}$82.8B tokens) \\
\quad OL (Option Level only)         & 1 epoch (${\approx}$108.7B tokens) \\
\quad Combined (FA+OL)               & 1 epoch (191.43B tokens) \\
\quad Cosmopedia-v2 (indiv.\ ablation) & 4 epochs (${\approx}$109.7B tokens) \\
\quad Cosmopedia-v2 (7 ep, matched budget)    & 7 epochs (${\approx}$192.4B tokens) \\
\multicolumn{2}{l}{\textit{Wall-clock time (64\,$\times$\,H100)}} \\
\quad FA (Failures only)             & 13\,h\,29\,min \\
\quad OL (Option Level only)         & 17\,h\,45\,min \\
\quad Combined (FA+OL)               & 31\,h\,31\,min \\
\quad Cosmopedia-v2 (indiv.\ ablation) & ${\sim}$17\,h \\
\quad Cosmopedia-v2 (7 ep, matched budget)    & ${\sim}$29.8\,h \\
\multicolumn{2}{l}{\textit{GPU-hours (64\,$\times$\,H100)}} \\
\quad FA (Failures only)             & ${\sim}$864 \\
\quad OL (Option Level only)         & ${\sim}$1,136 \\
\quad Combined (FA+OL)               & ${\sim}$2,017 \\
\quad Cosmopedia-v2 (indiv.\ ablation) & ${\sim}$1,088 \\
\quad Cosmopedia-v2 (7 ep, matched budget)    & ${\sim}$1,907 \\
\bottomrule
\end{tabular}
\end{table}

\begin{table}[h]
\caption{Benchmark contamination: QVAC Genesis III vs.\ Cosmopedia-v2.}
\label{tab:decon}
\centering
\small
\begin{tabular}{lrr}
\toprule
& \textbf{QVAC Genesis III} & \textbf{Cosmopedia-v2} \\
\midrule
Total documents          & 159.6M & 39.1M \\
Verified contamination   & 17     & 313   \\
GSM8K test leaked        & 0\%    & 0\%   \\
MMLU test leaked         & 0.005\% & 0.002\% \\
GSM8K train leaked       & 0.19\% & 3.52\% \\
\bottomrule
\end{tabular}
\end{table}

\section{Additional experiments and robustness}
\label{app:additional_results}

This section reports additional controls covering weak-student selection, a same-stage Cosmopedia-v2 comparison, transfer to non-Qwen architectures, parser sensitivity, and the separate parser failure rates underlying VAR.

\subsection{Weak-student model selection}

We selected the pretrained Qwen3-1.7B-Base model as the weak student used during data generation based on the controlled STEM evaluation in Table~\ref{tab:student_selection}, rather than on a model-family or leaderboard preference. This pretrained data-generation student is distinct from the randomly initialized checkpoints used in the controlled pretraining experiments.

\begin{table}[h]
\caption{Pretrained base-model comparison used to select the weak student for synthetic-data generation. All values are percentages under the same evaluation setup.}
\label{tab:student_selection}
\centering
\small
\begin{tabular}{lccccc}
\toprule
\textbf{Base model} & \textbf{MMLU avg} & \textbf{MMLU VAR} & \textbf{ARC-C} & \textbf{ARC-E} & \textbf{GPQA} \\
\midrule
Qwen3-1.7B-Base & \textbf{58.04} & \textbf{98.97} & \textbf{80.34} & \textbf{88.71} & \textbf{30.81} \\
SmolLM2-1.7B    & 41.11 & 95.81 & 58.64 & 77.60 & 23.74 \\
Llama-3.2-1B    & 29.96 & 96.94 & 32.54 & 48.15 & 23.74 \\
Gemma-3-1B-pt   & 14.52 & 55.88 & 12.88 & 19.05 & 15.15 \\
\bottomrule
\end{tabular}
\end{table}

\subsection{Same-stage comparison}

Table~\ref{tab:matched_checkpoint} compares the closest available checkpoints at \texttt{iter\_0003000}, corresponding to approximately 25B tokens for Cosmopedia-v2. Both QVAC splits substantially outperform Cosmopedia-v2 at the same training stage. Relative to Cosmopedia-v2, FA improves MMLU by 13.65 points and OL by 20.80 points; OL also improves ARC-C, ARC-E, and GPQA Diamond by 15.93, 16.94, and 8.08 points, respectively.

\begin{table}[h]
\caption{Same-stage comparison at \texttt{iter\_0003000}. All values are percentages.}
\label{tab:matched_checkpoint}
\centering
\small
\begin{tabular}{lccccc}
\toprule
\textbf{Model / training data} & \textbf{MMLU avg} & \textbf{MMLU VAR} & \textbf{ARC-C} & \textbf{ARC-E} & \textbf{GPQA} \\
\midrule
Cosmopedia-v2             & 8.98  & 37.59 & 13.90 & 15.34 & 12.63 \\
QVAC Genesis III FA       & 22.63 & 85.17 & 25.08 & 30.69 & 20.71 \\
QVAC Genesis III Combined & 22.91 & 80.63 & 23.05 & 27.34 & 18.18 \\
QVAC Genesis III OL       & \textbf{29.78} & \textbf{98.21} & \textbf{29.83} & \textbf{32.28} & \textbf{20.71} \\
\bottomrule
\end{tabular}
\end{table}

\subsection{Transfer across model architectures}

To test whether the OL benefit is specific to Qwen, we repeated the Cosmopedia-v2 4-epoch and QVAC-OL comparison with Llama-3.2-1B, SmolLM2-1.7B, and Gemma3-1B, using the same evaluation protocol. As shown in Table~\ref{tab:architecture_transfer}, QVAC-OL improves every reported metric for all three non-Qwen backbones. This transfer check is limited to OL, the strongest individual split.

\begin{table}[h]
\caption{Architecture-transfer results. Each QVAC-OL run is compared with the Cosmopedia-v2 4-epoch run using the same backbone and evaluation protocol. All values are percentages.}
\label{tab:architecture_transfer}
\centering
\small
\adjustbox{max width=\linewidth}{%
\begin{tabular}{llccccc}
\toprule
\textbf{Backbone} & \textbf{Training data} & \textbf{MMLU avg} & \textbf{MMLU VAR} & \textbf{ARC-C} & \textbf{ARC-E} & \textbf{GPQA} \\
\midrule
Qwen3-1.7B    & Cosmopedia-v2 (4 ep) & 20.39 & 72.71 & 21.02 & 20.63 & 17.67 \\
               & QVAC-OL               & 30.26 & 99.45 & 36.27 & 47.44 & 27.78 \\
\addlinespace
Llama-3.2-1B  & Cosmopedia-v2 (4 ep) & 14.42 & 51.65 & 20.00 & 21.69 & 18.18 \\
               & QVAC-OL               & 26.83 & 97.83 & 24.41 & 30.34 & 24.24 \\
\addlinespace
SmolLM2-1.7B  & Cosmopedia-v2 (4 ep) & 21.24 & 69.85 & 22.37 & 21.52 & 11.11 \\
               & QVAC-OL               & 34.36 & 99.78 & 32.54 & 41.98 & 26.26 \\
\addlinespace
Gemma3-1B     & Cosmopedia-v2 (4 ep) & 20.48 & 81.83 & 25.76 & 26.98 & 18.69 \\
               & QVAC-OL               & \textbf{34.71} & 99.66 & \textbf{48.81} & \textbf{62.96} & 24.75 \\
\bottomrule
\end{tabular}%
}
\end{table}

\subsection{Parser robustness and failure modes}

We re-evaluated the saved predictions from two representative endpoints with GPT-OSS-20B and GPT-OSS-120B as independent extractors. Table~\ref{tab:parser_robustness} shows that QVAC-OL's VAR advantage remains 24.28--36.97 points across all three extractors. The extractor only recovers the model's committed option; accuracy is computed afterward against the benchmark gold label.

\begin{table}[h]
\caption{Sensitivity to the answer-extraction model. Each cell reports MMLU accuracy / VAR (\%).}
\label{tab:parser_robustness}
\centering
\small
\begin{tabular}{lccc}
\toprule
& \multicolumn{3}{c}{\textbf{MMLU accuracy / VAR}} \\
\cmidrule(lr){2-4}
\textbf{Model} & \textbf{CompassJudger} & \textbf{GPT-OSS-20B} & \textbf{GPT-OSS-120B} \\
\midrule
Cosmopedia-v2 (7 ep) & 15.16 / 62.48 & 16.00 / 73.85 & 16.27 / 72.50 \\
QVAC Genesis III OL  & 30.26 / 99.45 & 29.71 / 98.13 & 30.23 / 99.31 \\
\bottomrule
\end{tabular}
\end{table}

Table~\ref{tab:parser_failure_rates} separates the two failure modes in Equation~\ref{eq:var}. QVAC-OL has a 0.06\% conflicting-answer rate, compared with 6.60\% and 16.90\% for the 4- and 7-epoch Cosmopedia-v2 baselines.

\begin{table}[h]
\caption{No-answer (\textsc{NoAns}), conflicting-answer (\textsc{Multi}), and valid-answer rates on MMLU. All values are percentages.}
\label{tab:parser_failure_rates}
\centering
\small
\begin{tabular}{lccc}
\toprule
\textbf{Model} & \textbf{MMLU NoAns} & \textbf{MMLU Multi} & \textbf{MMLU VAR} \\
\midrule
QVAC OL                  & 0.49  & 0.06  & 99.45 \\
QVAC Combined            & 6.80  & 1.14  & 92.06 \\
QVAC FA                  & 15.69 & 6.25  & 78.14 \\
Cosmopedia-v2 (4 ep)     & 20.69 & 6.60  & 72.71 \\
Cosmopedia-v2 (7 ep)     & 20.62 & 16.90 & 62.48 \\
\bottomrule
\end{tabular}
\end{table}

\section{Representative synthetic-data traces}
\label{app:data_examples}

The following compact examples trace one failure and one success from source seed to the resulting training document. \texttt{[...]} marks truncated text.

\subsection{Failure Analysis example}

\begin{promptbox}[title=1. Source seed (verbatim excerpt)]
\begin{promptverbatim}
The full moon on 23rd July 2021 – Friday, might seem
uncanny orange, especially to those who reside under
smoky skies prevailing from the burning wildfire
between summers.
[...]
Moreover, NASA also revealed that this full moon is
named Buck Moon after taking into account the Old Maine
Farmer’s Almanac 1930s publication about the full moon.
\end{promptverbatim}
\end{promptbox}

\begin{promptbox}[title=2. Generated MCQ]
\begin{promptverbatim}
What is the traditional name for July's full moon in
some cultures?
A) Harvest Moon
B) Strawberry Moon
C) Wolf Moon
D) Buck Moon
Gold answer: D
\end{promptverbatim}
\end{promptbox}

\begin{promptbox}[title=3. Weak-student response (verbatim excerpt)]
\begin{promptverbatim}
A

[...]

Question: What is the traditional name for July's full
moon in some cultures?

A. A) Harvest Moon
B. B) Strawberry Moon
C. C) Wolf Moon
D. D) Buck Moon

Answer: A
\end{promptverbatim}
\end{promptbox}

\begin{promptbox}[title=4. Parser output (verbatim excerpt)]
\begin{promptverbatim}
Extraction Reasoning: The candidate explicitly states
"Answer: A" directly after restating the original
question about July's full moon, indicating their final
choice.

Extracted Candidate's Answer: A
\end{promptverbatim}
\end{promptbox}

The extracted answer A differs from the gold answer D, so this record is routed to Failure Analysis.

\begin{promptbox}[title=5. Generated FA document (verbatim excerpt)]
\begin{promptverbatim}
# Understanding Lunar Naming Traditions and
Problem-Solving Strategies

## Problem Statement
**Question:** What is the traditional name for July’s
full moon in some cultures?
**Options:**
A) Harvest Moon
B) Strawberry Moon
C) Wolf Moon
D) Buck Moon

**Proposed Solution:** A (Harvest Moon)
**Correct Answer:** D (Buck Moon)

[...]

## Analysis of the Proposed Solution: Why "Harvest Moon"
(A) Is Incorrect

### Key Errors in the Proposed Answer
The proposed solution incorrectly identifies July’s full
moon as the **Harvest Moon**. This mistake likely arises
from two common misconceptions:
1. **Confusion with Autumn Timing**: The Harvest Moon
occurs in late summer/early autumn, not mid-summer
(July).
2. **Overemphasis on "Harvest"**: The term "harvest" is
strongly linked to agricultural labor, which peaks in
autumn, not July in temperate regions.

[...]

**Final Answer**
\boxed{D}
\end{promptverbatim}
\end{promptbox}

\subsection{Option-Level example}

\begin{promptbox}[title=1. Source seed (verbatim excerpt)]
\begin{promptverbatim}
Previous abstract Next abstract
Session 46 - Supernova Remnants.
Display session, Tuesday, January 14
We present an analysis of the X-ray data collected
during an observation of the supernova remnant (SNR)
MSH 11-62 by the Advanced Satellite for Cosmology and
Astrophysics (ASCA\/). We show that MSH 11-62 is a
composite remnant whose X-ray emission comes from two
distinct contributions: a nonthermal, synchrotron
emission, dominating the total flux above 2 keV, and
localized to a region of radius 3^\prime (consistent
with a point source) and a thermal component, extended
up to a radius of \sim5^\prime and detected only at
energies below 2 keV.
[...]
\end{promptverbatim}
\end{promptbox}

\begin{promptbox}[title=2. Generated MCQ]
\begin{promptverbatim}
What type of X-ray emission is typically associated
with synchrotron radiation in supernova remnants?
A. Thermal bremsstrahlung
B. Nonthermal synchrotron
C. Blackbody radiation
D. Gamma-ray bursts
Gold answer: B
\end{promptverbatim}
\end{promptbox}

\begin{promptbox}[title=3. Weak-student response (verbatim excerpt)]
\begin{promptverbatim}
B. Nonthermal synchrotron

The correct answer is B. Nonthermal synchrotron.

Explanation:
Synchrotron radiation is a type of electromagnetic
radiation produced when charged particles, such as
electrons, are accelerated to high energies in a
magnetic field. In the context of supernova remnants,
synchrotron radiation is typically associated with
nonthermal emission, which means it is produced by
non-thermal processes such as the acceleration of
particles in a magnetic field.

[...]
\end{promptverbatim}
\end{promptbox}

\begin{promptbox}[title=4. Parser output (verbatim excerpt)]
\begin{promptverbatim}
Extraction Reasoning: The candidate explicitly states
"The correct answer is B. Nonthermal synchrotron" and
begins their response with "B. Nonthermal synchrotron,"
indicating a clear final choice without any corrections
or conflicting statements.

Extracted Candidate's Answer: B
\end{promptverbatim}
\end{promptbox}

The extracted answer B matches the gold answer, so this record is routed to Option-Level Reasoning.

\begin{promptbox}[title=5. Generated OL document (verbatim excerpt)]
\begin{promptverbatim}
User: Hey, I’ve got this question here about X-ray
emissions in supernova remnants. The options are
A. Thermal bremsstrahlung, B. Nonthermal synchrotron,
C. Blackbody radiation, and D. Gamma-ray bursts. I’m a
bit stuck—could you walk me through it?

Assistant: Of course! Let’s start with the question:
*What type of X-ray emission is typically associated
with synchrotron radiation in supernova remnants?*
Hmm, okay. First, why is this question important?

[...]

User: Oh right, synchrotron radiation happens when
charged particles, like electrons, accelerate in
magnetic fields. I think it’s nonthermal because it’s
not from hot gas but from high-energy particles. But
how does that connect to supernova remnants?

Assistant: Nice! Supernova remnants (SNRs) are filled
with superheated gas and strong magnetic fields. When
shock waves from the supernova blast outward, they
energize electrons to nearly light-speed. These
electrons spiral around magnetic field lines, emitting
synchrotron radiation. Now, the question is about the
*type* of X-ray emission linked to this process.

[...]

User: That’s super helpful. So the answer is definitely
B?

Assistant: Yes! \boxed{B}

[...]
\end{promptverbatim}
\end{promptbox}

\section{Prompt templates}
\label{app:prompts}

This appendix documents the prompt templates used in our synthetic data pipeline. Placeholders are denoted with double braces (e.g., \texttt{\{\{level\}\}}) and are populated at runtime. Unless otherwise stated, templates are applied independently per MCQ instance.

\subsection{Scaling QA template (MCQ generation)}
\label{app:scaling_qa}

We generate MCQs from seed passages using a scaling QA template that enforces (i) self-contained questions, (ii) exactly four answer options, and (iii) an even distribution of correct choices across A/B/C/D.

\begin{promptbox}[title=Scaling QA template]
\begin{promptverbatim}
You are a {{level}} {{domain}} tutor. I will give you a context passage.
IGNORE the details of the passage. Use it only as inspiration
for topics, but DO NOT refer to the passage, book, text, or
"discussed/mentioned" material in your questions.
Context:
{{output}}

Generate EXACTLY 4 independent, self-contained questions suitable for {{level}} level {{domain}}.

Output rules (MUST be followed strictly):
- Output EXACTLY 4 samples.
- Each row = ONE question in the following format:
  "<question>","<choiceA>","<choiceB>",
  "<choiceC>","<choiceD>","<correct_choice_letter>"
- Each row MUST end with a newline (line break).
- Do NOT join multiple questions into one line.
- Do NOT output any extra text, headers, or blank lines.

Field definitions:
- <question> = a fully self-contained {{level}}-level {{domain}} problem.
  It must NOT reference any passage, book, figure, table,
  or "as discussed" or "as mentioned" style wording.
  Include all necessary values, constants, or definitions directly inside the question.
- <choiceA> ... <choiceD> = four mutually exclusive, plausible answer choices.
- <correct_choice_letter> = exactly one of A, B, C, D.

Question variety:
- Use a mix of imperative (e.g., "Calculate..."),
  interrogative, completion/fill-in, and true/false style.
- Do NOT start all questions with the same word.
- Include realistic values/constants when needed (e.g., g = 9.8 m/s²).
- Use standard SI units and symbols.

Hard constraints:
- Generate exactly 4 questions (4 samples).
- Correct answers must be evenly distributed: one A, one B, one C, one D.
- Randomize placement of correct answers; no clustering.
- Each row must contain exactly 6 comma-separated values.
- Do NOT add quotes around the entire block; only around individual fields.

Output ONLY the 4 samples in the required format, nothing else.
\end{promptverbatim}
\end{promptbox}

\subsection{MCQ answering template (student model prompting)}
\label{app:mcq_answer}

The answering model receives a plain MCQ prompt with a constrained answer field.

\begin{promptbox}[title=MCQ answering template]
\begin{promptverbatim}
Question: {{question}}

A. {{option_a}}
B. {{option_b}}
C. {{option_c}}
D. {{option_d}}

Answer:
\end{promptverbatim}
\end{promptbox}

\subsection{LLM-as-a-parser answer extraction template}
\label{app:parser}

We use an answer-extraction prompt to map free-form model outputs to a discrete option label, or to abstain when the response is ambiguous.

\begin{promptbox}[title=LLM-as-a-parser answer extraction template]
\begin{promptverbatim}
MMLU_ANSWER_EXTRACTOR_TEMPLATE = """
You are an expert answer extractor. Your ONLY job is to
extract the final answer from the candidate's response.

CRITICAL INSTRUCTIONS - READ CAREFULLY:
- DO NOT solve the question yourself
- DO NOT generate a new answer
- DO NOT think about what the correct answer should be
- DO NOT evaluate whether the candidate's answer is right or wrong
- ONLY extract what the candidate actually wrote as their final answer

Your task is purely extraction, not generation or evaluation.

Here are the extraction guidelines for MMLU multiple choice questions:
1. Look for the candidate's final answer in their response.
   This should be one of: A, B, C, or D
   - Look for explicit statements like "ANSWER: A", "The answer is B", "Final answer: C"
   - Look for \boxed{A} format (extract what's inside the braces)
   - Look for standalone letters A, B, C, or D in their conclusion
   - The final choice they settle on in their reasoning

2. If the candidate's response contains multiple different answers:
   - If they state one answer but then correct/fix themselves, extract the corrected/final answer they provided
   - If they state multiple different answers without
     choosing one or correcting themselves,
     return "MULTIPLE_ANSWERS"

3. If you cannot clearly identify a single letter choice
   (A, B, C, or D) in the response, return "NO_ANSWER"

4. If the candidate generates new questions:
   - DO NOT extract the answer from the new generated questions
   - ONLY extract the answer from the original question

RESPONSE FORMAT:
First, provide a brief explanation of why you are
extracting that particular answer (what indicators you
found in the candidate's response).
Then, provide the extracted answer as a single letter.

Use this exact format:

Extraction Reasoning: [Brief explanation of what
indicators led you to extract this answer from the
candidate's response]

Extracted Candidate's Answer: [A single letter:
A, B, C, or D. Use MULTIPLE_ANSWERS if the candidate
provided multiple different answers, or NO_ANSWER if
no clear answer was found.]

<Original Question Begin>:
Question: {input}

A. {A}
B. {B}
C. {C}
D. {D}

Answer:
<Original Question End>

<Candidate's Response Begin>:
{prediction}
<Candidate's Response End>
"""
\end{promptverbatim}
\end{promptbox}

\subsection{Failure Analysis (FA) rendering templates}
\label{app:fa_templates}

For failed instances, we render corrective explanations in four styles. Each template receives the original problem (\texttt{\{\{prompt\}\}}), the model's proposed solution (\texttt{\{\{full\_response\}\}}), and the gold target (\texttt{\{\{target\}\}}). All outputs are required to be self-contained and to place the final answer in \texttt{\textbackslash boxed\{\dots\}}.

\begin{promptbox}[title=Educational textbook style (FA)]
\begin{promptverbatim}
You are an educational content creator generating high-quality textbook explanations for academic learning.

Given:
  The problem: {{prompt}}
  A proposed solution: {{full_response}}
  The correct answer: {{target}}

Content Requirements:
- Generate a comprehensive educational content of MAXIMUM 3000 words
- Final answers must appear within \boxed{...}
- Write in clear, pedagogical language suitable for textbooks
- Create appropriate section titles and structure that fit the specific topic and problem type
- Organize your explanation with logical sections that help students understand the concept,
  identify errors in the proposed solution, learn the correct approach, and apply these insights

Your Task:
Analyze the given problem and proposed solution. Create a comprehensive educational explanation that:
1. Includes the complete problem statement clearly within your explanation
2. Introduces the key concepts and principles relevant to this problem type
3. Examines the proposed solution, identifying where and why it goes wrong
4. Provides the correct solution approach that leads to the given target answer
5. Explains the underlying principles that make the correct method work
6. Discusses broader applications and common misconceptions
7. Concludes with actionable takeaways for students

IMPORTANT: Your textbook explanation must be completely self-contained. Include the full problem statement
within your response so readers have all necessary information without needing external context.

Structure your response with appropriate headings and sections that naturally fit the subject matter and problem type
\end{promptverbatim}
\end{promptbox}

\begin{promptbox}[title=Web article style (FA)]
\begin{promptverbatim}
You are a content creator specializing in engaging, informative web articles that break down complex problems and solutions.

Given:
  The problem: {{prompt}}
  A proposed solution: {{full_response}}
  The correct answer: {{target}}

Content Requirements:
- Generate engaging web content of MAXIMUM 3000 words
- Use a conversational yet informative tone suitable for online readers
- Final answers must appear within \boxed{...}
- Create compelling headings and subheadings that work well for web reading
- Include relatable examples and practical insights
- Structure content for easy scanning with shorter paragraphs and clear sections

Your Task:
Create an engaging web article that breaks down this problem and solution. Your article should:
1. Start with the complete problem statement presented in an engaging way
2. Hook readers by explaining why this type of problem matters in real life
3. Analyze the proposed solution, showing where it goes wrong in an accessible way
4. Walk through the correct solution step-by-step with clear explanations
5. Explain why the correct approach works using relatable analogies when helpful
6. Share practical tips and common pitfalls readers should watch out for
7. End with actionable takeaways and encourage further learning

IMPORTANT: Your article must be completely self-contained and include the full problem statement. Write for a general
audience interested in learning. Use engaging language that makes complex concepts accessible.

Structure your response with compelling headings that would work well for web content and encourage readers to keep reading.
\end{promptverbatim}
\end{promptbox}

\begin{promptbox}[title=Conversational dialogue style (FA)]
\begin{promptverbatim}
You are creating a natural conversational dialogue between a curious student and a knowledgeable assistant discussing a problem and its solution.

Given:
  The problem: {{prompt}}
  A proposed solution: {{full_response}}
  The correct answer: {{target}}

Content Requirements:
- Generate a comprehensive natural conversational dialogue of MAXIMUM 3000 words
- Use "User:" and "Assistant:" to clearly mark each speaker
- Final answers must appear within \boxed{...}
- Make the conversation flow naturally with realistic student questions
- Include follow-up questions and clarifications that feel authentic
- Create an engaging back-and-forth that teaches through dialogue

Your Task:
Create a natural conversation where a student asks about this problem and you provide helpful explanations. The dialogue should:
1. Start with the student presenting the complete problem they're working on
2. Include the student asking why this type of problem is important or interesting
3. Have the student share their attempted solution and ask for feedback
4. Show the assistant explaining what went wrong in a supportive way
5. Include the student asking for the correct approach step-by-step
6. Have natural follow-up questions about the reasoning and methods
7. End with the student asking for tips to avoid similar mistakes in the future

IMPORTANT: Create a completely self-contained dialogue that includes the full problem statement naturally within the conversation.
Make it feel like an authentic tutoring session with realistic questions and responses.

Present the entire response as a natural dialogue using "User:" and "Assistant:" labels.
\end{promptverbatim}
\end{promptbox}

\begin{promptbox}[title=Question--answer style (FA)]
\begin{promptverbatim}
You are an expert tutor providing clear, direct answers to questions about problem-solving and reasoning.

Given:
  The problem: {{prompt}}
  A proposed solution: {{full_response}}
  The correct answer: {{target}}

Content Requirements:
- Generate a focused Q&A content of MAXIMUM 3000 words
- Use clear, direct language with a helpful tutoring tone
- Final answers must appear within \boxed{...}
- Structure as natural Q&A flow that addresses the key learning points
- Focus on practical understanding and clear explanations
- Prioritize clarity and directness over lengthy explanations

Your Task:
Create a focused Q&A response that addresses this problem and solution. Your response should:
1. Present the complete problem clearly as the main question
2. Identify what makes this type of problem important or challenging
3. Analyze what went wrong in the proposed solution with specific examples
4. Provide the correct solution with step-by-step reasoning
5. Explain the key principle or method that ensures the right approach
6. Give practical advice for avoiding similar mistakes
7. Summarize the main takeaway in a clear, memorable way

IMPORTANT: Your Q&A must be completely self-contained and include the full problem statement. Write as if directly answering a
student's question, focusing on the most important insights they need to understand.

Structure your response in a natural Q&A format that flows logically from question to comprehensive answer.
\end{promptverbatim}
\end{promptbox}

\subsection{Option-Level Reasoning (OL) rendering templates}
\label{app:ol_templates}

For successful instances, OL produces contrastive analyses in four styles. Each template receives the original MCQ (\texttt{\{\{prompt\}\}}) and the gold target (\texttt{\{\{target\}\}}). All outputs must be self-contained and must analyze the correct option first, followed by each incorrect option.

\begin{promptbox}[title=Educational textbook style (OL)]
\begin{promptverbatim}
You are an educational content creator generating high-quality textbook explanations for multiple choice question analysis.

Given:
  The question: {{prompt}}
  The correct answer: {{target}}

Content Requirements:
- Generate MAXIMUM 3000 words of educational content
- Final answers must appear within \boxed{...}
- Write in clear, pedagogical language suitable for textbooks
- Create appropriate section titles and structure that fit the specific topic and problem type
- Organize your explanation with logical sections that help students understand the concept, analyze the correct option first,
  then the incorrect ones

Your Task:
Analyze the given multiple choice question and all answer options. Create a comprehensive educational explanation that:
1. Includes the complete question and all options clearly within your explanation
2. Introduces the key concepts and principles relevant to this problem type
3. First analyzes the correct answer option in detail, explaining why it is right with step-by-step reasoning
4. Provides thorough verification of why the correct option is accurate, identifying key principles and logical reasoning
5. Then examines each incorrect answer option systematically, explaining why each is wrong
6. Identifies specific errors, misconceptions, or flaws in each incorrect option
7. Discusses the underlying principles that distinguish correct from incorrect reasoning
8. Explores broader applications and common misconceptions related to this topic
9. Concludes with actionable strategies for approaching similar questions

IMPORTANT: Your textbook explanation must be completely self-contained. Include the full question and all answer options within
your response so readers have all necessary information without needing external context. Always analyze the correct option first,
then the incorrect ones.

Structure your response with appropriate headings and sections that naturally fit the subject matter and question type, focusing on
systematic analysis starting with the correct option.
\end{promptverbatim}
\end{promptbox}

\begin{promptbox}[title=Web article style (OL)]
\begin{promptverbatim}
You are a content creator specializing in engaging, informative web articles that break down multiple choice questions and effective analysis strategies.

Given:
  The question: {{prompt}}
  The correct answer: {{target}}

Content Requirements:
- Generate MAXIMUM 3000 words of engaging web content
- Use a conversational yet informative tone suitable for online readers
- Final answers must appear within \boxed{...}
- Create compelling headings and subheadings that work well for web reading
- Include relatable examples and practical insights
- Structure content for easy scanning with shorter paragraphs and clear sections

Your Task:
Create an engaging web article that breaks down this multiple choice question and demonstrates effective analysis. Your article should:
1. Start with the complete question and all options presented in an engaging way
2. Hook readers by explaining why mastering multiple choice analysis matters in real life
3. First analyze the correct answer option in detail, walking through why it's right with clear explanations
4. Provide step-by-step verification of the correct option using relatable analogies when helpful
5. Then systematically analyze each incorrect answer option, explaining why each is wrong
6. Show the specific errors or misconceptions in each incorrect option using accessible language
7. Share practical tips and common pitfalls readers should watch out for when tackling similar questions
8. End with actionable takeaways and strategies for improving multiple choice performance

IMPORTANT: Your article must be completely self-contained and include the full question and all answer options. Write for a general
audience interested in learning better test-taking and analytical thinking skills. Use engaging language that makes complex reasoning
accessible. Always analyze the correct option first, then the incorrect ones.

Structure your response with compelling headings that would work well for web content and encourage readers to keep reading through the
analysis of the correct option first, then the incorrect ones.
\end{promptverbatim}
\end{promptbox}

\begin{promptbox}[title=Question--answer style (OL)]
\begin{promptverbatim}
You are an expert tutor providing clear, direct answers about multiple choice question analysis and reasoning.

Given:
  The question: {{prompt}}
  The correct answer: {{target}}

Content Requirements:
- Generate MAXIMUM 3000 words of focused Q&A content
- Use clear, direct language with a helpful tutoring tone
- Final answers must appear within \boxed{...}
- Structure as natural Q&A flow that addresses the key learning points
- Focus on practical understanding and clear explanations
- Prioritize clarity and directness over lengthy explanations

Your Task:
Create a focused Q&A response that addresses this multiple choice question. Your response should:
1. Present the complete question and all options clearly as the main question
2. Identify what makes this type of question important or challenging
3. First analyze the correct answer option, explaining why it is right with step-by-step logic
4. Provide clear verification of the correct option's reasoning and principles
5. Then analyze each incorrect answer option systematically, explaining why each is wrong
6. Give specific examples of the errors or misconceptions in each incorrect option
7. Share practical advice for approaching similar multiple choice questions
8. Summarize the key principle or method that helps distinguish correct from incorrect options

IMPORTANT: Your Q&A must be completely self-contained and include the full question and all answer options. Write as if directly
answering a student's question about how to analyze multiple choice options effectively. Always analyze the correct option first, then
the incorrect ones.

Structure your response in a natural Q&A format that flows logically from the question to analysis of the correct option, then the incorrect options.
\end{promptverbatim}
\end{promptbox}

\begin{promptbox}[title=Conversational dialogue style (OL)]
\begin{promptverbatim}
You are creating a natural conversational dialogue between a curious student and a knowledgeable assistant discussing a multiple choice question and its analysis.

Given:
  The question: {{prompt}}
  The correct answer: {{target}}

Content Requirements:
- Generate MAXIMUM 3000 words of natural conversational dialogue
- Use "User:" and "Assistant:" to clearly mark each speaker
- Final answers must appear within \boxed{...}
- Make the conversation flow naturally with realistic student questions
- Include follow-up questions and clarifications that feel authentic
- Create an engaging back-and-forth that teaches multiple choice analysis through dialogue

Your Task:
Create a natural conversation where a student asks about this multiple choice question and you provide helpful explanations. The dialogue should:
1. Start with the student presenting the complete question and all options they're working on
2. Include the student asking why this type of question is important or how to approach it
3. Have the student share their initial thoughts about the options and ask for guidance
4. Show the assistant first explaining why the correct option is right with detailed step-by-step reasoning
5. Include the student asking for clarification about the correct option's reasoning
6. Have the assistant then analyze each incorrect option, explaining why each is wrong
7. End with the student asking for general tips to improve at multiple choice questions

IMPORTANT: Create a completely self-contained dialogue that includes the full question and all answer options naturally within the conversation.
Make it feel like an authentic tutoring session with realistic questions and responses about multiple choice strategy. Always analyze the correct option
first, then the incorrect ones.

Present the entire response as a natural dialogue using "User:" and "Assistant:" labels.
\end{promptverbatim}
\end{promptbox}

\end{document}